\def\year{2018}\relax
\documentclass[letterpaper]{article} 
\usepackage{aaai18}  
\usepackage{times}  
\usepackage{helvet}  
\usepackage{courier}  
\usepackage{url}  
\usepackage{graphicx}  
\usepackage{latexsym}
\usepackage{mwe}
\usepackage{xcolor}
\usepackage{multirow}
\usepackage{microtype}
\usepackage{lipsum}
\usepackage{multibib}
\usepackage{pbox}
\usepackage{epstopdf}
\usepackage{ragged2e}
\usepackage{array}
\usepackage{url}
\usepackage{booktabs}
\usepackage{relsize}
\usepackage[shortlabels]{enumitem}
\usepackage[fleqn]{amsmath}
\usepackage{makecell}
\usepackage{tabularx}
\usepackage{placeins}
\usepackage{xhfill}
\usepackage{makecell}
\usepackage{ragged2e}
\newcommand*{\dittoclosing}{--''--}
\makeatletter
\newcommand{\@BIBLABEL}{\@emptybiblabel}
\newcommand{\@emptybiblabel}[1]{}
\makeatother

\begin{document}
%

\title{Complex Sequential Question Answering: Towards Learning to Converse Over Linked Question Answer Pairs with a Knowledge Graph}
\author{Amrita Saha\textsuperscript{1,2}\thanks{Equal Contribution} \qquad Vardaan Pahuja \textsuperscript{3}\footnotemark[1]\thanks{Work done while at IBM Research} \qquad\qquad Mitesh M. Khapra\textsuperscript{2} \\ 
\texttt{amrsaha4@in.ibm.com} \qquad \texttt{vardaanpahuja@gmail.com} \qquad \texttt{miteshk@cse.iitm.ac.in} \\
\\
  {\bf \Large Karthik Sankaranarayanan\textsuperscript{1} \qquad Sarath Chandar\textsuperscript{3}} \\
  \qquad\qquad\qquad \texttt{kartsank@in.ibm.com}  \qquad \texttt{apsarathchandar@gmail.com} \\
    $^1$\Large{IBM Research AI} \qquad
    $^2$\Large{I.I.T. Madras, India}\qquad
    $^3$\Large{MILA, Universit\'e de Montr\'eal}
}  
\maketitle
\begin{abstract}
While conversing with chatbots, humans typically tend to ask many questions, a significant portion of which can be answered by referring to large-scale knowledge graphs (KG). While Question Answering (QA) and dialog systems have been studied independently, there is a need to study them closely to evaluate such real-world scenarios faced by bots involving both these tasks. Towards this end, we introduce the task of Complex Sequential QA which combines the two tasks of (i) answering factual questions through complex inferencing over a realistic-sized KG of millions of entities, and (ii) learning to converse through a series of coherently linked QA pairs. Through a labor intensive semi-automatic process, involving in-house and crowdsourced workers, we created a dataset containing around 200K dialogs with a total of 1.6M turns. Further, unlike existing large scale QA datasets which contain simple questions that can be answered from a single tuple, the questions in our dialogs require a larger subgraph of the KG. Specifically, our dataset has questions which require logical, quantitative, and comparative reasoning as well as their combinations. This calls for models which can: (i) parse complex natural language questions, (ii) use conversation context to resolve coreferences and ellipsis in utterances, (iii) ask for clarifications for ambiguous queries, and finally (iv) retrieve relevant subgraphs of the KG to answer such questions. However, our experiments with a combination of state of the art dialog and QA models show that they clearly do not achieve the above objectives and are inadequate for dealing with such complex real world settings. We believe that this new dataset coupled with the limitations of existing models as reported in this paper should encourage further research in Complex Sequential QA. 
\end{abstract}

\section{Introduction}

In recent years there has been an increased demand for AI driven personal assistants which are capable of conversing coherently with humans. Such personal assistants could benefit from large scale knowledge graphs which contain millions of facts stored as tuples of the form \{\textit{predicate, subject, object}\} (for example, \{\textit{director, Titanic, James Cameron}\}). Such knowledge graphs can indeed be handy when the bot is used in specific domains such as education, entertainment, sports, etc. where it is often required to answer factual questions while being aware of the context of the conversation. While Question Answering \cite{voorhees2000building,wang2007jeopardy,DBLP:conf/emnlp/YangYM15,berant2013semantic,DBLP:journals/corr/BordesUCW15,rajpurkar2016squad,DBLP:journals/corr/NguyenRSGTMD16,onishi2016did,DBLP:conf/emnlp/RichardsonBR13,DBLP:conf/emnlp/BerantSCLHHCM14} and Conversation Systems \cite{DBLP:conf/naacl/RitterCD10,DBLP:conf/sigdial/LowePSP15,DBLP:conf/acl/Banchs12,DBLP:journals/corr/BordesW16} have received a lot of attention in the recent past, we would like to focus on such real life settings encountered by chatbots which involve a combination of QA and dialog. Specifically, we are interested in building systems which can learn to converse over a series of coherently linked questions that can be answered from a large scale knowledge graph. We refer to this task as Complex Sequential Question Answering (CSQA).

\begin{table}[!ht]
    \centering
    \scriptsize{
   \setlength{\tabcolsep}{0.1em} 
    {
    \begin{tabular}{|p{0.52cm}p{1.7cm}p{5.7cm}|}\hline
        \textbf{Turn} & \textbf{~~~State} & \textbf{Utterance} \\\hline
        \multirow{2}{*}{\textbf{T1}}&\textbf{Simple Q.} & \textbf{USER :} Can you tell me which cities border Verderio Inferiore ? \\
        &\textbf{Response} & \textbf{SYSTEM :} Cornate d'Adda, Bernareggio, Robbiate \\ \hline
        \multirow{2}{*}{\textbf{T2}}&\textbf{Coreference Q.} & \textbf{USER :} And which cities flank that one ? \\
        &\textbf{Clarification Q.} & \textbf{SYSTEM :} Did you mean Robbiate ?\\\hline
        \multirow{2}{*}{\textbf{T3}}&\textbf{Clarification} & \textbf{USER :}  No, I meant Cornate d'Adda. \\
        &\textbf{Response} & \textbf{SYSTEM :} Bottanuco, Busnago, Trezzo sull'Adda\\\hline
        \multirow{2}{*}{\textbf{T4}}&\textbf{Logical Q.} & \textbf{USER :} Which ones share border with Robbiate and are not part of Azerbaijan ? \\
        &\textbf{Response} & \textbf{SYSTEM :} Trezzo sull'Adda, Busnago, Bottanuco \\\hline
        \multirow{2}{*}{\textbf{T5}}&\textbf{Quantitative Q.} &\textbf{USER :} Which city is the capital of min number of french administrative divisions ?\\
        &\textbf{Response} & \textbf{SYSTEM :} Riga \\\hline
        \multirow{2}{*}{\textbf{T6}}&\textbf{Boolean Q.} & \textbf{USER :} Is Rot am See located adjacent to that one ? \\
        &\textbf{Response} & \textbf{SYSTEM :} No \\\hline
        \multirow{2}{*}{\textbf{T7}}&\textbf{Quanti. Q.} & \textbf{USER :}  How many political territories have strategic relation with around 22 other countries ? \\
        &\textbf{Response} &\textbf{SYSTEM :} 11 \\\hline
        \multirow{2}{*}{\textbf{T8}}&\textbf{Logical Q.} &\textbf{USER :} Which are the nations connected with Robbiate and are not located in Italy ? \\
        &\textbf{Response} &\textbf{SYSTEM :} The answer count is 11. Do you want to see all possibilities?  \\\hline
        \multirow{2}{*}{\textbf{T9}}&\textbf{Clarification} &\textbf{USER :}  No, show only a few of them \\
        &\textbf{Response} &\textbf{SYSTEM :} Papua New Guinea, Jamaica, Antigua and Barbuda, Austria, New Zealand, The Bahamas, Gabon \\\hline
        \multirow{2}{*}{\textbf{T10}}&\textbf{Quantitative Q.} &\textbf{USER :} How many countries are diplomatically related to Italy ? \\
        &\textbf{Response} &\textbf{SYSTEM :} 74 \\\hline
        \multirow{2}{*}{\textbf{T11}}&\textbf{Quantitative Q.} &\textbf{USER :} And how many of them also with Australia ? \\
        &\textbf{Response} &\textbf{SYSTEM :} 56 \\\hline
         \multirow{2}{*}{\textbf{T12}}&\textbf{Comparative Q.} &\textbf{USER :} Which countries have lesser number of political relations with others than Australia ? \\
        &\textbf{Response} &\textbf{SYSTEM :} Denmark, Canada, Grenada \\\hline
    \end{tabular}
    }
    }
    \caption{A sample dialog from the dataset (More examples of generated dialogs are provided in the supplementary material)}
    \label{tab:dialogue_example}
\end{table}

Needless to say, CSQA is very different from the kind of conversations found in existing dialog datasets such as the Twitter \cite{DBLP:conf/naacl/RitterCD10}, Ubuntu \cite{DBLP:conf/sigdial/LowePSP15} and Movie Subtitles \cite{DBLP:conf/acl/Banchs12} datasets. Table~\ref{tab:dialogue_example} shows an example of one such conversation from our dataset containing a series of questions. Note that to answer the question in Turn 11, the bot needs to remember that the question involves the same predicate (`diplomatically related') as the previous question, but with a different subject (`Australia'). In other words, it is difficult to answer this question without retaining the context of the conversation. Further, in a natural conversation, some of the questions may require co-reference resolution (as in Turn 2), ellipsis resolution (as in Turn 11), etc. Finally, in some cases the question could be ambiguous (as in Turn 2) in which case the bot needs to ask for clarifications keeping in mind other entities and relations which were previously mentioned in the conversation.

While the example in Table~\ref{tab:dialogue_example} already highlights some of the challenges involved in CSQA, we now discuss an orthogonal set of challenges which arise from the complexity of the questions. Existing datasets for Factual QA \cite{berant2013semantic,DBLP:journals/corr/BordesUCW15,DBLP:conf/emnlp/YangYM15} deal with \textit{Simple Questions}, each of which can be answered from a single tuple in the KG. However, in a real-life setting, a bot could encounter more complex questions requiring logical, quantitative and comparative reasoning. Table \ref{tab:question_types} shows some examples of such questions. It should be clear that unlike simple questions, which can be answered from a single tuple, these questions require a larger subgraph of the KG. For example, to answer the question ``Which rivers flow through India and China ?" one needs to find (i) the set of rivers flowing through India (ii) the set of rivers flowing through China and finally (iii) the intersection between these two sets. Answering such questions requires models which can parse complex natural language questions, retrieve relevant subgraphs of the KG and then perform some logical, comparative and/or quantitative operations on this subgraph. Also the Knowledge Graph used in our work is orders of magnitude larger than those used in some existing works \cite{DBLP:journals/corr/BordesW16,DBLP:journals/corr/BordesUCW15,DBLP:journals/corr/DodgeGZBCMSW15,DBLP:conf/emnlp/FaderSE11} which lie at the intersection of QA and dialog.

Having motivated the task of CSQA and highlighted its differences from existing work on dialog and QA, we now briefly describe the process used for creating our dataset. As mentioned earlier, a KG contains tuples of the form \{\textit{predicate, subject, object}\}. For each of the 330 predicates in our Knowledge Graph, we first asked workers on Amazon Mechanical Turk to create questions containing the predicate and the subject (or object) such that the answer to that question is the object (or subject). These questions are complete in the sense that they do not have any ambiguity and can be answered in isolation without requiring any additional context. We then ask in-house annotators to create multiple templates for generating a conversation comprising of connected question answer pairs. Two question answer pairs are said to be connected if they contain the same predicate, subject or object. We then ask workers to make modifications to these questions so as to introduce challenges like co-references, ellipsis, incompleteness (or under specification) and contextual dependence. We also solicit templates and modifications to add logical, comparative and quantitative operators to the questions obtained above. This results in a dataset which contains conversations of the form shown in Table \ref{tab:dialogue_example}. 

The objective of this work is twofold: (i) to introduce the task of Complex Sequential QA and (ii) to show the inadequacy of current state of the art QA and dialog methods to deal with such tasks. Towards the second objective, we propose a model for CSQA which is a cross between a state of the art hierarchical conversation model \cite{Serban:2016:BED:3016387.3016435} and a key value based memory network model for QA \cite{DBLP:journals/corr/MillerFDKBW16}. Through our experiments, we demonstrate the inadequacy of these models and highlight specific challenges that need to be addressed. It is also worth mentioning that the unambiguous (context independent) questions which appear in our dataset (typically, at the start of the conversation) can also be used for studying Complex Question Answering (as opposed to Simple Question Answering) in isolation ignoring the dialog context. This will help to independently push the state of the art in Complex QA.

\section{Related Work}
Our work lies at the intersection of Question Answering and Dialog Systems. Question Answering has always been of interest to the research community starting from early TREC evaluations \cite{voorhees2000building} . Over the years various datasets and tasks have been introduced to advance the state of the art in QA. These datasets can be divided into 5 main types (i) TREC style Open Domain QA \cite{voorhees2000building,wang2007jeopardy,DBLP:conf/emnlp/YangYM15} where the aim is to answer a question from a collection of documents (ii) factoid QA over structured knowledge graphs \cite{berant2013semantic,DBLP:journals/corr/BordesUCW15,DBLP:conf/acl/SerbanGGACCB16} (iii) reading comprehension style QA \cite{rajpurkar2016squad,DBLP:journals/corr/NguyenRSGTMD16}, (iv) cloze style QA \cite{DBLP:journals/corr/MostafazadehCHP16,onishi2016did} (v) multiple choice QA \cite{DBLP:conf/emnlp/RichardsonBR13,DBLP:conf/emnlp/BerantSCLHHCM14}

Of the above QA tasks, factoid QA is most relevant to us as the questions in our CSQA dataset are factoid questions. Existing factoid QA datasets contain \textit{Simple Questions} which can be answered from a single tuple in the knowledge graph. Specifically, unlike our dataset, none of the existing datasets contain \textit{complex questions} requiring logical, quantitative and comparative reasoning involving larger subgraphs of the KG as opposed to a single tuple. Solutions to the simple QA task range from semantic parsing based methods 
\cite{berant2014semantic,fader2014open}  to embedding based methods \cite{DBLP:conf/pkdd/BordesWU14,bordes2014question,yang2014joint} and state of the art Memory Networks based architectures \cite{DBLP:journals/corr/BordesUCW15,DBLP:journals/corr/MillerFDKBW16,DBLP:conf/icml/KumarIOIBGZPS16}. In this work, we experiment with Memory Network based architectures and make a case for the need of better architectures when going beyond simple questions. 

Since we are interested in CSQA which contains a series of QA pairs over a coherent conversation, we also review some related work on dialog systems. Over the past few years three large scale dialog datasets, \textit{viz.}, Twitter-Conversations \cite{DBLP:conf/naacl/RitterCD10}, Ubuntu Dialogue \cite{DBLP:journals/dad/LowePSCLP17} and  Movie-Dic Corpus \cite{DBLP:conf/acl/Banchs12} have become very popular. However, none of these datasets have the flavor of CSQA and there is no explicit Knowledge Graph associated with the conversations. Here again, neural network based (hierarchical) sequence to sequence methods \cite{luong2015multi,Serban:2016:BED:3016387.3016435,serban2017hierarchical} have become the de facto choice. Recently, \cite{DBLP:journals/corr/BordesW16} proposed a dataset which contains knowledge graph driven goal oriented dialogs for the task of restaurant reservation. However, the size of the KG here is very small (\textless 10 cuisines, locations, ambience, etc.) and the dialog contains very few states. \cite{DBLP:journals/corr/DodgeGZBCMSW15} also uses a dataset for QA and recommendation but unlike our dataset, their dataset does not contain coherently linked question answer pairs. Further, the KG is again much smaller (75K entities and 11 relations). Recently \cite{DBLP:journals/corr/NeelakantanLAMA16} have explored complex question answering over around 18.5K queries from the WikiTableQuestions dataset, but their tables have less than 100 rows and a handful of columns whereas our complex QAs are grounded in a KB of over 20 million tuples. Further, their dataset does not have a conversational aspect. 

\section{Dataset Creation}
\label{sec:dataset}

Our aim is to create a dataset which contains a series of linked QA pairs forming a coherent conversation. Further, these questions should be answerable from a Knowledge Graph using logical, comparative and/or quantitative reasoning. We started by asking pairs of in-house annotators to converse with each other. One annotator in the pair acted as a \textit{user} whose job was to ask questions and the other annotator acted as the \textit{system} whose job was to answer the questions or ask for clarifications if required. Note that these annotators were Computer Science graduates who understood the concepts of knowledge graph, sub graph, tuples, subject, object, relation, etc. The idea was to use the in-house annotators to understand the types of simple and complex questions that can be asked over a knowledge graph. These could then be abstracted to templates and used to instantiate more questions involving different relations, subjects and objects. Similarly, we also wanted to understand the type of coreferences, ellipses etc used by users when asking linked questions over a coherent conversation. These could again be abstracted to templates and used to link individual QA pairs to form a coherent dialog. In the remainder of this section we describe (i) the knowledge graph supporting our CSQA (ii) simple question templates suggested by the in-house annotators (iii) complex question templates and finally (iv) the linked conversation templates and the process used to instantiate around 200K dialogs containing 1.6 million linked QA pairs. 

\subsection{Knowledge Graph}
As our KG, we used wikidata which stores facts in the form of tuples containing a relation, a subject and an object. For example, \emph{(rel: capital, subj: India, obj: New Delhi)} is a tuple in wikidata. Each entity (subject or object) is associated with an entity type. For example, in the above tuple, \textit{India} is an entity of type \textit{country} and \textit{New Delhi} is an entity of type \textit{city}. We use the wikidata dump of 14-Nov-2016 which contains 5.2K relations, 12.8M entities and 52.3M facts. 
Of these 5.2K relations, we retain only 330 meaningful ones. Specifically, we ignore relations such as ``ISO 3166-1 alpha-2 code'', ``NDL Auth ID'' etc., as we do not expect users to ask questions about such obscure relations. 
Similarly, of the 30.8K unique entity types in wikidata, we selected 642 types (considering only immediate parents of entities) which appeared in the top 90 percentile of the tuples associated with atleast one of the retained meaningful relations. In effect, there were around 21.2M  such tuples containing only the filtered relations and entity types. The total number of unique entities in these filtered tuples is 12.8M out of which 3.8M appear in atleast 3 tuples. 

\begin{table}[!ht]
{\scriptsize
\begin{center}
{
{\setlength{\tabcolsep}{0.6em}
\begin{tabular}{|p{0.69cm}|p{1.5cm}|p{1.4cm}|p{3.3cm}|} \hline
	\textbf{Reaso- ning} & \textbf{Type} & \textbf{Containing} &  \textbf{Example} \\ \hline
	\multirow{5}{0.69cm}{Log- ical} & Union & \multirow{4}{1.4cm}{\begin{tabular}[c]{@{}l@{}}Single\\Relation\end{tabular}} &  Which rivers flow through India or China?\\ \cline{2-2}\cline{4-4}
	 & Intersection &  &  Which rivers flow through India and China?\\ \cline{2-2}\cline{4-4}
	 & Difference &  & Which rivers flow through India but not China?\\ \cline{2-4}
	 & Any of the above & Multiple Relations &   Which river flows through India but does not originate in Himalayas?\\ \hline
	Verifi-cation & Boolean & Single/Multi- ple entities &  Does Ganga flow through  India ? \\ \hline
	\multirow{9}{0.69cm}{Quant- itative} & \multirow{2}{1.5cm}{Count} & Single entity type &    How many rivers flow through  India ?\\ \cline{3-4}
	 &  & Mult. entity type  &  How many rivers and lakes does India have ?\\ \cline{3-4}
	 &  & \begin{tabular}[c]{@{}l@{}}Logical\\ operators\end{tabular} & How many rivers flow through India and/or/but not China?\\ \cline{2-4}	 
	 & \multirow{2}{1.5cm}{Min/Max} & Single entity type & Which river flows through maximum number of countries ?\\ \cline{3-4}
	 &  & Mult. entity type  &  Which country has maximum number of rivers and lakes combined ?\\ \cline{2-4}
	 & Atleast\slash Atmost & Single entity type  &   Which rivers flow through at least N countries ?\\ \cline{3-4}
	 & \slash Approx.\slash Equal & Mult. entity type &  Which country has at least N rivers and lakes combined ?\\ \cline{2-4}
	 & \multirow{2}{1.5cm}{Count over Atleast  } & Single entity type & How many rivers flow through at least N countries?\\  \cline{3-4}
	 &  \slash Atmost \slash Approx./Equal & Mult. entity type &   How many countries have at least N rivers and lakes combined ?\\\hline
	\multirow{4}{0.69cm}{Comp- arative} & \multirow{2}{1.5cm}{More\slash Less} & Single entity type &   Which countries have more number of rivers than India ?\\ \cline{3-4}
	 &  & Mult. entity type & Which countries have more rivers and lakes than India ?\\ \cline{2-4}
	 & \multirow{2}{1.5cm}{Count over More\slash Less} & Single entity type &   How many countries have more number of rivers than India ?\\ \cline{3-4}
	 &  & Mult. entity type &  How many countries have more rivers and lakes than India ?\\ \hline
\end{tabular}
}
}
\end{center}
}
\caption{Types of questions in the dataset.}
\label{tab:question_types}
\end{table}

\subsection{Simple Questions}
For discovering simple question templates, we asked the annotators to come up with questions which can be answered from a single tuple in the knowledge graph. The annotators suggested that for a given tuple (say, \textit{rel: CEO, subj: Google, obj: Sundar Pichai}) there are mainly 3 types of simple questions that can be generated:
\vspace{0.2cm}

\textbf{1. Object based questions:} Here the question contains the relation and the subject from a tuple and the answer is the tuple's object. For example, ``\textbf{Q:} Who is the \textit{CEO} (relation) of \textit{Google} (subject) ? \textbf{A:} \textit{Sundar Pichai} (object)''.

\textbf{2. Subject based questions:} Here the question contains the relation and the object from a tuple and the answer is the tuple's subject. For example, ``\textbf{Q:} Which company is \textit{Sundar Pichai} (object) the \textit{CEO} of (relation) ? \textbf{A:} \textit{Google} (subject)''.

\textbf{3. Relation based questions:} Here the question contains the subject and the object from a tuple and the answer is the tuple's relation. For example, ``\textbf{Q:} How is \textit{Sundar Pichai} (object) related to \textit{Google} (subject) ? \textbf{A:} \textit{CEO} (relation)''. During our discussions, we found that in many cases, relation based questions do not make a lot of sense. For example, it is unnatural for someone to ask the question ``\textbf{Q:} How is Himalayas related to India? \textbf{A:} located in''. 
Hence, in this work we focus only on object based and subject based questions. 
\vspace{0.1cm}

Note that in some cases the question could have multiple correct answers. In other words, there are multiple tuples related to this question.  For example, ``\textbf{Q:} Which rivers flow through India ? \textbf{A:}
Ganga, Yamuna, Narmada, ....''. Note that even though these questions can be answered from multiple tuples, they are still simple questions because they do not require any joint reasoning over multiple tuples. 
\vspace{0.2cm}

\textbf{Crowdsourced question generation:} Based on this initial pilot with in-house annotators we then requested workers on AMT to create subject based and object based questions for each of the 330 relations in our KG. For creating subject based questions the annotators were shown (i) the object, (ii) the relation (iii) the type of the subject associated with that tuple and (iv) a few sample tuples. 
Note that the subject type is important as the annotator will need to look at the subject type (city) to form the question ``Which \emph{city} is the capital of India ?''. This is important because some relations (for example, the relation \textit{tributary}) can have multiple subject and object types as shown below:
\begin{enumerate}
\item \textit{subj: Spring Creek (type: river), obj: Lake Ilsanjo (type: lake)}
\item \textit{subj: Spring Creek (type: river), obj: Matanzas Creek (type: stream)}
\end{enumerate}
It should be obvious that even for the same relation different combinations of subjects and objects should result in different questions. For example, ``Which \textit{lake} is a tributary of Spring Creek?'' v/s ``Which \textit{river} is a tributary of Spring Creek?''. Note that, on an average each relation in our KG was associated with 5 subject types and 6 object types. We first asked a set of workers to create one subject based and object based question for each relation. We then asked a separate set of annotators to create paraphrases of these questions. In all, we collected 1531 subject based and 1450 object based question templates (including paraphrases) through this process. Once we get a template we can instantiate it with different entity types and entities to create many questions. For example, given the template ``Which $<$\textit{water\_course}$>$ is located in $<$\textit{country}$>$ ?'' we can instantiate it by replacing \textit{water\_course} by it's sub-types (river, lake, etc) and by replacing \textit{country} by entities of that type (U.S., India, etc.). This gives us a semi-automatic way of creating many questions from the collected templates. Note that the question templates also contain paraphrases, so we have different ways of asking the same question.

\subsection{Complex Questions}
Next we wanted the annotators to help us identify types of questions which require logical, comparative and quantitative reasoning over a larger subgraph of the KG. 
\vspace{0.2cm}

\textbf{Logical Reasoning:} These are questions which require some logical inferencing over multiple tuples in the KG. For example, consider the question ``Which rivers flow through India and China ?'' To answer this question we first need to create two sets (i) a set $A$ containing rivers appearing in tuples of the form \textit{(flows through, India, river)} and (ii) a set $B$ containing rivers appearing in tuples of the form \textit{(flows through, China, river)}. The final answer to the question is then an intersection of these two sets. It should be obvious that answering such questions is more difficult then the \textit{Simple Questions} studied in literature so far (and as described in the previous section). 

The annotators came up with questions involving different logical operators such as AND, OR, NOT, etc (see  Table \ref{tab:question_types}). They also suggested some templates for creating such logical reasoning questions from the simple questions that we had already collected (as described in Section 3.1). For example, one such template was to take a simple object based question such as ``Which rivers flow through India'' and augment it with another subject such as ``and China''.  Similar templates and paraphrases were suggested for other operators such as OR, NOT, etc. for both subject based and object based questions. This allowed us to semi-automatically create many questions requiring logical reasoning. This process is semi-automatic because once a template is created, we instantiate it for multiple tuples (as explained earlier) and then manually verify a subset of these questions to check whether they are syntactically and semantically correct. 

Note that some of the logical reasoning questions suggested by the annotators contained multiple relations. For example, the question ``Which river flows through India and has its source in Himalayas?'' requires a logical operation over two relations, \textit{viz., flows\_through} and \textit{source}.

\vspace{0.2cm}

\textbf{Quantitative Reasoning:}
These questions require some quantitative reasoning involving standard aggregation functions like max, min, count, atleast / atmost / approximately / equal to $N$, etc. We refer the reader to Table \ref{tab:question_types} to see examples of different types of quantitative questions. 
Once again, with the help of in-house annotators we identified several templates for modifying the simple questions that we had already collected and creating quantitative reasoning questions involving different aggregation operators. For example, one such template was to take the object based question ``Which rivers flow through India'' and replace ``Which'' by ``How many''. In fact, we found this particular template to be so convenient that for every relation, we asked the workers to give us at least one simple question which starts with ``Which \emph{subject-type} ... ''. Some of these simple questions starting with ``Which \emph{subject-type} ... '' look a bit unnatural but we made a conscious choice to allow this so that it simplifies the process of creating complex questions. We also created questions which require quantitative reasoning on top of logical reasoning. For example, ``How many rivers flow through India but not through China ?''. 
\vspace{0.2cm}

\textbf{Comparative Reasoning:}
These are questions which require a comparison between entities based on certain relations (predicates). 
For example, consider the question ``Which countries have more number of rivers than India ?''. This requires inference over multiple tuples in the KG. The model here essentially needs to learn the count, sort and more/less operations. 
Such questions could also involve multiple entity types. For example the question ``Which countries have more lakes and rivers than India ?'' involves two entity types (lakes, rivers). 
Finally, we could have questions which require a counting type quantitative reasoning on top of comparative reasoning. For example, ``How many countries have more rivers than India ?'' requires counting after comparing. These questions were created by modifying the simple questions, using the rules of transformation given by our annotators.


Note that in all of the above cases, once the annotator suggests a modification, we can apply that modification and its paraphrases to multiple tuples to get many questions. Further, after instantiating we retain only those Qs which have less than 1000 answers.
\vspace{0.2cm}

\subsection{Linked Sequential QA}
So far we have described the process of collecting individual QA pairs containing various types of questions. We are now interested in creating coherent conversations involving such QA pairs. We can think of such a conversation as a walk over the Knowledge Graph using QA pairs such that subsequent questions refer to subjects, objects or relations which have appeared previously in the conversation. More specifically, such conversations should have the following properties (i) subsequent QA pairs should be linked and (ii) the conversation should contain typical elements of a dialog such as coreferences, ellipses, clarifications, confirmation, etc. 

The process of connecting linked QA pairs in a coherent conversation can be thought of as performing a systematic walk over a Knowledge Graph. Simply stated, two questions can be placed next to each other in a conversation if they share a relation or an entity. However, bringing in factors such as ambiguity, underspecified or coreferenced questions into the conversation requires manual effort. For this, we again requested in-house annotators to create templates for converting simple or complex questions described above into conversational questions. For example, one such template was to take a simple question such as ``Which rivers flow through India ?'' and replace the subject by ``that \emph{subject-type}'' or ``that country'' in this case. Multiple such templates were created and refined for different question types that we described in the earlier sections. This was a labor intensive tedious process requiring several iterations. Some templates were also collected using crowdsourcing on AMT. We refer to such questions as Indirect questions as opposed to Direct questions which are fully specified and do not indirectly refer to some entity or relation from the earlier conversation.  
The in-house annotators also suggested some clarification templates which involved asking questions containing coreferences which could resolve to more than one of the previously mentioned entities. Turn 2 in Table \ref{tab:dialogue_example} shows one such example. The information in this question is not enough to answer the question and hence the system needs to ask for a clarification. Note that, whenever we use linking we only link consecutive questions and not arbitrary questions in the sequence (\textit{i.e.}, the $i$-th question can be linked to the next pr previous question but not to arbitrary questions appearing before or after it.) 

Through the above processing involving a mix of manual work (crowdsourced and inhouse) and semi-automatic instantiation, we created a dataset containing 200 K dialogs and a total of 1.6 M turns. Table \ref{tab:question_types} shows the number of templates for each question type and some sample types. Table ~\ref{tab:dataset_stats} shows various statistics about the dataset including the Train, Validation and Test splits 
. Note that we constructed the train, valid and test splits in such a way that the dialogs in the validation and test set do not contain questions corresponding to tuples for which questions were seen at train time.

\begin{table}[!ht]
{\scriptsize
\begin{center}
\begin{tabular}{|p{4.32cm}|p{0.73cm}|p{0.62cm}|p{0.62cm}|}\hline
\textbf{Dataset Statistics} & \textbf{Train} & \textbf{Valid} &  \textbf{Test} \\ \hline
Total No. of Dialogs(chat sessions) & 152391 & 16413 & 27797 \\\hline
Avg. No. of Utterances per dialog & 15.9 & 15.65 & 19.44\\\hline
Total No. of Utterances having Question/Answer & 1.2M  & .13M & .27M\\\hline
Length of user's question (in words) & 9.7 & 9.68 & 10.28\\\hline
Length of system's response (in words) & 4.74 & 4.67 & 4.37\\\hline
Avg. No. of Dialog states per dialog & 3.89 & 3.84 & 4.53\\\hline
Vocab size (freq\textgreater=10) & 0.1M & -  & - \\\hline
\end{tabular}
\end{center}
}
\caption{Overall Dataset Statistics}
\label{tab:dataset_stats}
\end{table}

\section{Some peculiar characteristics of Wikidata} 
We found that Wikidata has some typical characteristics and predicates, subject types and object types which often leads to very unnatural questions. We list down some of these issues below:

\begin{itemize}
    \item \textbf{Very generic predicates}: Consider the relation $lake\_outflow$ for which the annotators suggested the question ``Which $object\_type$ outflows from the lake YYY ?''. This seems like a valid template but turns out that Wikidata also contains predicates of the form $lake\_outflow(Dal Lake, evaporation)$ where evaporation is an outflow from the lake. Similarly, the relation $fabrication\_method$ allows for methods used to \textit{grow, cook, weave, build, assemble, manufacture an item}. Due to the presence of such very generic relations (which allow a wide range of object types) sometime the questions instantiated from these templates may look very unnatural. In many cases, we manually tried to filter out such questions but given the scale of the KB it was not always possible to do this. We expect some such noisy questions to be a part of the final dataset. 
    
    \item \textbf{Overlapping predicate and subject types}: The word \textit{religion} is both a predicate and a subject type in Wikidata. Similarly, \textit{sport} is both a predicate and a subject type in Wikidata. This often leads to some questions containing repitions (for example, ``Which religion (subject type) is the religion (predicate) practised by YYY ?''. Again, we filtered out many such cases by applying some rule based post-processing after instantiating the templates but we still expect a few of these to be present in the dataset. 
    
    \item \textbf{Long tail of subject types and relations}: There are  a few subject types in Wikidata which are very dominant. For example, a large number of entities in Wikidata belong to the sub-class \textit{person} and \textit{location}. These subject types in turn are associated with a few dominant relations. For example, \textit{part-of} is the predominant relations associated with almost entities of type \textit{location}. Similarly, \textit{citizen-of, birthdate, birthplace} are common relations associated with almost all entities of type \textit{person}. Other relations such as \textit{named-after} are a bit rare. Hence, when creating complex or linked questions connecting multiple entities and relations some of the rarer relations do not show up frequently. Such long tail behavior wherein some relations and predicates dominate will be observed in any KB of a reasonable size and can't really be avoided.
    
    \item \textbf{Unnatural Peer Subject types:} As per Wikidata, the subject types \textit{religion} and \textit{social group} are peers as they are both sub-classes of \textit{belief system}. As a consequence of this we have logical questions of the form ``Which religions and social groups does YYY belong to?''. We found this is a bit odd and we are not sure if an average user would consider these to be peers. These are special cases and are expected to any such large scale KB.

\end{itemize}

\section{Proposed Model}
\label{sec:models}

\begin{figure*}
\begin{center}
\includegraphics[width=\textwidth,height=6.6cm]{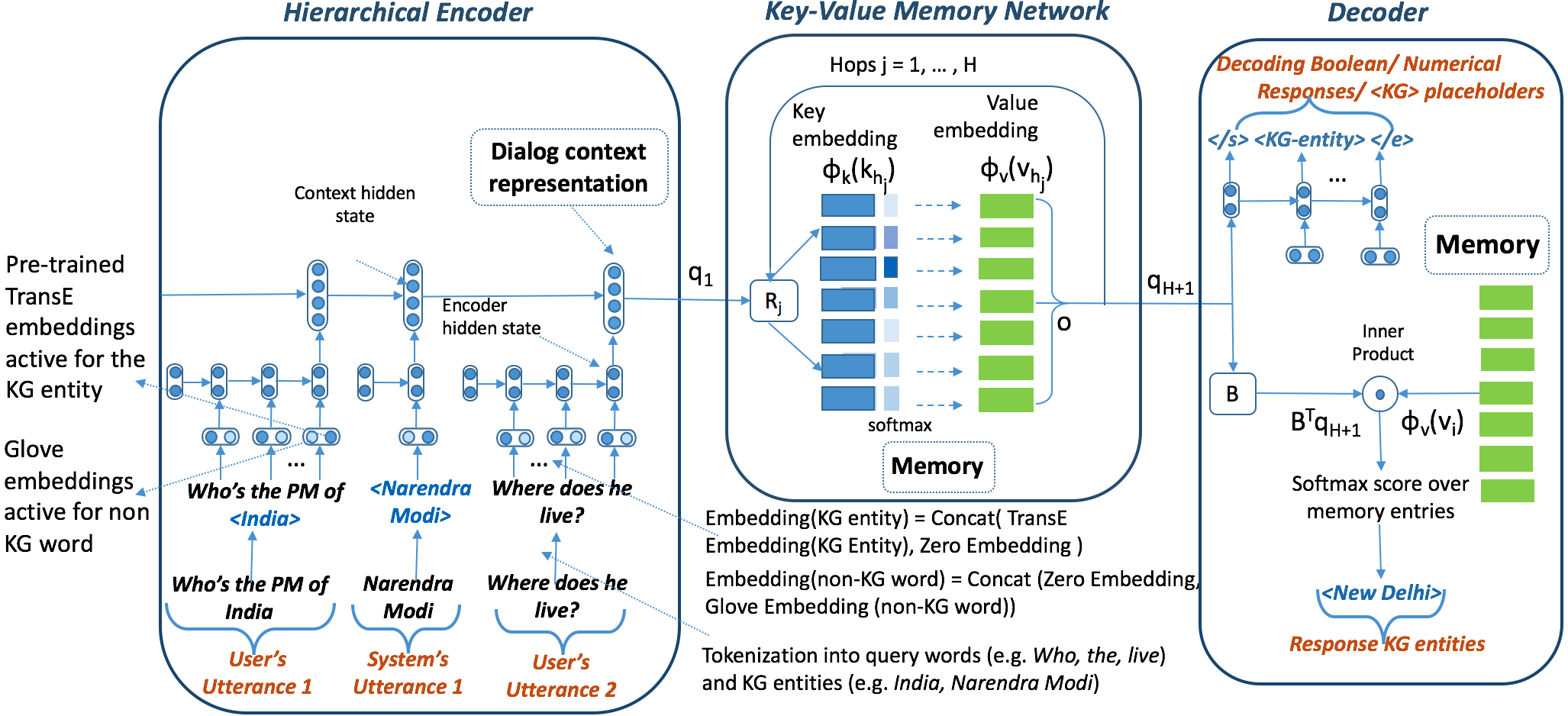}  
\end{center}
\caption{Proposed Model consisting of a (i) Hierarchical Encoder (ii) Key-Value Memory Network and (iii) Decoder }
\label{fig:model}
\end{figure*}

Since CSQA involves a combination of dialog and QA, we propose a model which is a cross between (i) the HRED model \cite{Serban:2016:BED:3016387.3016435} which is one of the state of the art models for dialog systems and (ii) the key value memory network model \cite{DBLP:journals/corr/MillerFDKBW16} which is a state of the art QA system. Our model has the following components:

\vspace{0.5em}
\noindent \textbf{1. Hierarchical Encoder:} The model contains a lower level RNN encoder which goes over the words in an utterance and computes a representation for each utterance. This is followed by a higher level encoder which goes over these utterance representations and computes a representation $q_1$ for the context (current state of the dialog). 

\vspace{0.5em}
\noindent \textbf{2. Handling Large Vocabulary:} As input to the above encoder, we provide pre-trained Glove embeddings \cite{pennington2014glove} of the words in the question. However, our questions contain many entities (names, locations, etc.) for which pre-trained word embeddings are not available. Since these entities are crucial for answering the questions we cannot treat them as unknown words. One option is to randomly initialize the embeddings of these entity words and then train them along with other parameters of the model. This would effectively lead to a very large vocabulary and blow up the number of parameters. To avoid this, we use a state of the art TransE method \cite{DBLP:conf/nips/BordesUGWY13} for learning embeddings of KG entities offline. More specifically, for entities such as India, China, Ganga, Himalayas, etc. which are present in the KG we learn embeddings using the TransE model. We refer to these embeddings as KG embeddings. The final embedding of every question word is then a concatenation of the Glove embedding (if available, 0s otherwise) and the KG embedding (if available, 0s otherwise). 

\vspace{0.5em}
\noindent \textbf{3. Candidate generation:} State of the art memory network based methods \cite{DBLP:journals/corr/MillerFDKBW16} learn to compute an attention function over the tuples in the KG based on the given question (or dialog context in our case). For large sized KGs, it is infeasible to compute the attention over the entire KG. Instead, following \cite{DBLP:journals/corr/MillerFDKBW16} we filter out tuples from the KG using the longest possible n-gram matching. We essentially consider only the longest n-gram which corresponds to the name of a KG-entity and retain only those tuples where the entity appears as  subject/object. We observed that even with this filtering, the average number of candidate tuples for a given question in our dataset can sometimes be very large. We return to this issue in the Discussions section. 

\vspace{0.5em}
\noindent \textbf{4. Key Value Memory Network:} A key value memory network stores each of the $N$ candidate tuples (as selected above) as a key-value pair where the key contains the concatenated embedding of the relation and the subject (denoted by $\phi_K(k_{h_i}) \in R^{D}$ for the $i^{th}$ memory entry) whereas the value contains the embedding of the object (denoted by $\phi_V(v_{h_i}) \in R^{D}$ for the $i^{th}$ memory entry). Here, the subject, object and relation embeddings are the TransE KG embeddings, as described above. The model makes multiple passes over the memory computing new attention weights over the keys of the memory at each pass and updating the contextual question representation ($q$) whose initial representation $q_1 \in R^{d}$ is computed by the hierarchical encoder. The rationale behind making multiple passes over the question is that the model may learn to focus on different aspects of the question in each pass. This is especially important in the case of complex questions. The following equation shows how the query representation gets updated in the $j^{th}$ pass.
{
\begin{equation}
\label{myeq:one}
    q_{j+1} = R_j (q + \sum_i^{N} \text{Softmax}(q_{j} A\phi_K(k_{h_i})) A\phi_V(v_{h_i}))
\end{equation}
}

$A \in R^{d\times D}$ and $R_{1\ldots H} \in R^{d\times d}$ are the parameters of the key-value memory network and $N$ is the number of candidate tuples.

\vspace{0.5em}
\noindent \textbf{5. Decoder:} For a truly end-to-end solution, the decoder should be generic enough to produce multiple types of answers. For example, here are some of the answer sequences that the decoder is expected to generate: (i) \textit{5 rivers and 4 lakes} (for count questions) (ii) \textit{Yes/No/Yes and No respectively etc.} (for verification questions) (iii) \textit{Did you mean ...} (for clarification questions) (iv) \textit{Ganga, Narmada, Yamuna, ...} (list of KG entities satisfying the query) and so on. At a high level, we can say that the model always produces sequences and in most cases the sequences will contain KG entities whereas in some cases the sequences may contain counts, entity types (rivers, lakes, etc) and non-KG words. We thus model the decoder as an RNN based sequence generator which takes as input the modified query representation. At each time step it gives a softmax over a shortlisted vocabulary containing counts, yes/no and KG entity types amounting to 1500 words approximately. Note that even though the model has to produce KG entities, we cannot include all KG entities in this vocabulary (as it will blow up the number of parameters). Instead, we train the decoder to produce the token $KG\_WORD$ whenever a KG entity needs to be produced in the output. We then use a copy mechanism to replace the $KG\_WORD$ with relevant entities. For example, if the decoder produces $n$ $KG\_WORD$ tokens then we use $q_{H+1}$ to give a distribution over the entities in the candidate tuples and then replace each $KG\_WORD$ token in the output by these top $n$ entities having the highest probability. The distribution over the candidate entities is computed as $\text{Softmax}( q_{H+1}B \phi_V(v_{h_i}))$ where $B \in R^{d\times D}$ is a parameter. The training loss is a sum of the cross-entropy loss over the tokens and the KG entities.

\begin{table}[!ht]
{\scriptsize
\centering
\begin{center}
{
{
\begin{tabular}{|p{4.5cm}||p{0.89cm}|p{0.94cm}|}\hline
\textbf{Question Type} & \emph{\textbf{Recall}} & \emph{\textbf{Precision}} \\\hline
\pbox{2.2cm}{Overall }& 15.83\% & 6.7\%\\ \hline
\pbox{4.2cm}{Simple Question (Direct)}&  27.9\% &7.77\%\\ \hline
\pbox{4.2cm}{Simple Question (Coreferenced)} & 12.31\% & 3.84\%  \\ \hline
\pbox{4.2cm}{Simple Question (Ellipsis)}& 19.45\% & 3.96\% \\ \hline
\pbox{4.2cm}{Logical Reasoning (All)}& 27.22\% & 10.52\% \\ \hline
\pbox{4.2cm}{Quantitive Reasoning (All)} & 0.29\% & 0.44\% \\ \hline
\pbox{4.5cm}{Comparative~Reasoning (All)} & 1.26\% & 5.45\%\\ \hline
\pbox{2.2cm}{Clarification} & 30.64\% & 10.8\% \\ \hline\hline
\textbf{Question Type} &  \multicolumn{2}{c|}{\emph{\textbf{F1 Score}}}\\ \hline
Verification (Boolean) (All) & \multicolumn{2}{c|}{17.68\%}  \\ \hline
Quantitative Reasoning (Count) (All)& \multicolumn{2}{c|}{40.2\%} \\ \hline
Comparing Reasoning (Count) (All)& \multicolumn{2}{c|}{11.86\%} \\ \hline\hline
\textbf{Question Type} &  \multicolumn{2}{c|}{\emph{\textbf{BLEU-4}}}\\ \hline
Clarification (Natural Language Generation) & \multicolumn{2}{c|}{15.58} \\ \hline
\end{tabular}
}}
\end{center}
}

\caption{Performance of the proposed model on different types of questions in the dialog}
\label{tab:results}
\end{table}

\section{Results}

We used Adam as the optimization algorithm and tuned the following hyperparameters using the validation set; learning rate $\in$ \{1e-3, 4e-4\}, RNN hidden unit size, word embedding size, KG embedding size $\in$ \{256, 512\}, batch size $\in$ \{32, 64\} and dialog context size as 2. The bracketed numbers indicate the values of each hyperparameter considered. On average, we found that the candidate generation step produces 10K candidate tuples, hence we kept upto 10K key value pairs in the memory network. Following \cite{DBLP:journals/corr/MillerFDKBW16}, we set $H=2$. We used Precision and Recall as the evaluation metrics which capture the percentage of entities in the final decoder output that were correct and the percentage of actual entities that were retrieved by the system respectively. For verification and count based questions which produce a sequence of Yes and/or No or counts we use accuracy as the evaluation metric (\textit{i.e.}, whether the count or boolean answer was exact or not). Finally for questions which need clarification, the system has to generate a natural language response which is usually a sequence of KG-entities and non-KG words, hence for that we separately report both Precision/Recall over the predicted KG-entities and BLEU for the overall utterance similarity. The results of our experiments are summarized in Table ~\ref{tab:results}.


\section{Discussions}
Based on the results in Table~\ref{tab:results}, we discuss some shortcomings of existing methods and suggest areas for future research.

\paragraph{\textbf{1. Simple v/s Complex Questions:}} It is obvious that the model performs very poorly on complex questions as compared to simple questions. There are multiple reasons for this. First, existing models do not really model an aggregate or logical function for handling quantitative, comparative and logical reasoning. Designing such aggregation functions for an end-to-end solution is non-trivial and needs further exploration. This dataset should provide a good benchmark for exploring such solutions for complex QA. Second, it is not clear if the existing encoders (HRED + KVmem, in this case) are capable of effectively parsing complex questions and feeding a good represetation to the decoder. For example, the encoder ideally needs to learn to break down the question ``Which rivers flow through India and China?'' into two parts (i) ``Which rivers flow through India?'' (ii) ``Which rivers flow through China?'' and then compute an attention over relevant tuples in the memory. Such kind of parsing is not explicitly modeled by existing encoders. There is clearly a need for revisiting some of the traditional parsing based methods for QA in the light of this dataset.

\paragraph{\textbf{2. Direct v/s Indirect Questions:}} Comparing the third and fourth rows of Table~\ref{tab:results} with the second row, we see that the performance of the model drops when dealing with indirect or incomplete questions which rely on the context for resolving ellipsis, coreferences, etc. Even though current dialog systems (HRED, in this case) do learn to capture the context, one key challenge w.r.t our dataset is that, here named entities and relations matter more than other words in the context. We need better models which can explicitly learn to give importance to relations and entities (for example, using an explicit supervised attention mechanism).

\paragraph{\textbf{3. Candidate Generation:}}This step is required to prune the size of the KG and store only relevant steps in the memory. This step is a bit adhoc as it relies on n-gram matching and we saw specific issues while using this on our dataset. We had explicitly asked the annotators to create paraphrases of the same question. As a result simple n-gram matching does not work well resulting in low recall of the actual answer entity in the filtered candidate tuples. A better candidate matching algorithm which takes care of entity paraphrases (Leo, Leonardo, etc.) and relation paraphrases (director, directed by, direct, etc.) are needed. In some cases, we also have the reverse problem. For example, if the entity being referred to in the question is extremely popular then it will be involved in over 100K tuples in the KB (for example, an entity like U.S.A.). This causes the KV memory to blow up leading to poor and inefficient training and inference. 

\paragraph{\textbf{4. Better organization of the memory:}} It is inevitable that for some questions, especially complex questions involving logical operators over multiple entities and relations, the number of tuples required to be stored in the memory would be large. For example, around 15\% of the questions in our data require more than 100K candidate tuples. Current Key Value Memory Networks which are flat in their organization are not suitable for this for two reasons. First, the amount of memory required by the model increases and can go beyond the capacity of existing GPUs. Second the attention weights computed using equation~\ref{myeq:one} need a prohibitively expensive softmax computation which increases both training and test time. Better ways of organizing the memory along with approximate methods for computing the softmax function are needed to handle such complex questions.

We hope that the dataset, results and discussions on the resources presented in this paper will convince the reader that CSQA has several challenges which are not encountered in previous datasets for dialog and QA. Some of them are listed above and there are a few more which we do not list due to space constraints. Addressing/solving all of these challenges is clearly beyond the scope of a single paper. The purpose of this paper was to introduce the task and propose a model based on existing state of the art models and thereby highlight the need for further research to address the inadequacies of these models. To facilitate research, this dataset will be made available at  \url{https://github.com/iitm-nlp-miteshk/AmritaSaha/tree/master/CSQA} (please copy paste the URL in a browser instead of clicking on it). This URL will contain the following resources:

\normalsize{}
\begin{itemize}
    \item the train/valid/test splits used in our experiments
    \item the processed version of the WikiData dump of 14-Nov-2016 that was used to construct the dataset
    \item scripts to extract the train/valid/test set for each of the different question types listed in Table \ref{tab:question_types} 
    \item scripts to evaluate the performance of the model
\end{itemize}

\section{Conclusion}
In this paper, we introduced the task of Complex Sequential Question Answering (CSQA) with a large scale dataset consisting of conversations over linked QA pairs. The dataset contains 200K dialogs with 1.6M turns and was collected through a manually intensive semi-automated process. To the best of our knowledge, this is the first dataset of its kind which contains complex questions which require logical, quantitative and/or comparative reasoning over a large Knowledge Graph containing millions of tuples. We propose a model for CSQA which is a cross between state of the art models for dialog and QA and highlight the inadequacies of this model in dealing with the task of CSQA. It should be obvious that CSQA has several challenges and addressing/solving all of them is beyond the scope of a single paper. We hope that the introduction of this task and dataset should excite the research community to develop models for Complex Sequential Question Answering.

\bibliography{aaai}

\begin{thebibliography}{}

\bibitem[\protect\citeauthoryear{Banchs}{2012}]{DBLP:conf/acl/Banchs12}
Banchs, R.~E.
\newblock 2012.
\newblock Movie-dic: a movie dialogue corpus for research and development.
\newblock In {\em ACL, 2012},  203--207.

\bibitem[\protect\citeauthoryear{Berant and Liang}{2014}]{berant2014semantic}
Berant, J., and Liang, P.
\newblock 2014.
\newblock Semantic parsing via paraphrasing.
\newblock In {\em ACL (1)},  1415--1425.

\bibitem[\protect\citeauthoryear{Berant \bgroup et al\mbox.\egroup
  }{2013}]{berant2013semantic}
Berant, J.; Chou, A.; Frostig, R.; and Liang, P.
\newblock 2013.
\newblock Semantic parsing on freebase from question-answer pairs.
\newblock In {\em EMNLP}, volume~2, ~6.

\bibitem[\protect\citeauthoryear{Berant \bgroup et al\mbox.\egroup
  }{2014}]{DBLP:conf/emnlp/BerantSCLHHCM14}
Berant, J.; Srikumar, V.; Chen, P.; Linden, A.~V.; Harding, B.; Huang, B.;
  Clark, P.; and Manning, C.~D.
\newblock 2014.
\newblock Modeling biological processes for reading comprehension.
\newblock In {\em {EMNLP} 2014,}.

\bibitem[\protect\citeauthoryear{Bordes and
  Weston}{2016}]{DBLP:journals/corr/BordesW16}
Bordes, A., and Weston, J.
\newblock 2016.
\newblock Learning end-to-end goal-oriented dialog.
\newblock {\em CoRR} abs/1605.07683.

\bibitem[\protect\citeauthoryear{Bordes \bgroup et al\mbox.\egroup
  }{2013}]{DBLP:conf/nips/BordesUGWY13}
Bordes, A.; Usunier, N.; Garc{\'{\i}}a{-}Dur{\'{a}}n, A.; Weston, J.; and
  Yakhnenko, O.
\newblock 2013.
\newblock Translating embeddings for modeling multi-relational data.
\newblock In {\em Neural Information Processing Systems 2013},  2787--2795.

\bibitem[\protect\citeauthoryear{Bordes \bgroup et al\mbox.\egroup
  }{2015}]{DBLP:journals/corr/BordesUCW15}
Bordes, A.; Usunier, N.; Chopra, S.; and Weston, J.
\newblock 2015.
\newblock Large-scale simple question answering with memory networks.
\newblock {\em CoRR} abs/1506.02075.

\bibitem[\protect\citeauthoryear{Bordes, Chopra, and
  Weston}{2014}]{bordes2014question}
Bordes, A.; Chopra, S.; and Weston, J.
\newblock 2014.
\newblock Question answering with subgraph embeddings.
\newblock {\em arXiv preprint arXiv:1406.3676}.

\bibitem[\protect\citeauthoryear{Bordes, Weston, and
  Usunier}{2014}]{DBLP:conf/pkdd/BordesWU14}
Bordes, A.; Weston, J.; and Usunier, N.
\newblock 2014.
\newblock Open question answering with weakly supervised embedding models.
\newblock In {\em {ECML} {PKDD} 2014. Proceedings, Part {I}},  165--180.

\bibitem[\protect\citeauthoryear{Dodge \bgroup et al\mbox.\egroup
  }{2015}]{DBLP:journals/corr/DodgeGZBCMSW15}
Dodge, J.; Gane, A.; Zhang, X.; Bordes, A.; Chopra, S.; Miller, A.~H.; Szlam,
  A.; and Weston, J.
\newblock 2015.
\newblock Evaluating prerequisite qualities for learning end-to-end dialog
  systems.
\newblock {\em CoRR} abs/1511.06931.

\bibitem[\protect\citeauthoryear{Fader, Soderland, and
  Etzioni}{2011}]{DBLP:conf/emnlp/FaderSE11}
Fader, A.; Soderland, S.; and Etzioni, O.
\newblock 2011.
\newblock Identifying relations for open information extraction.
\newblock In {\em {EMNLP} 2011},  1535--1545.

\bibitem[\protect\citeauthoryear{Fader, Zettlemoyer, and
  Etzioni}{2014}]{fader2014open}
Fader, A.; Zettlemoyer, L.; and Etzioni, O.
\newblock 2014.
\newblock Open question answering over curated and extracted knowledge bases.
\newblock In {\em Proceedings of the 20th ACM SIGKDD international conference
  on Knowledge discovery and data mining},  1156--1165.
\newblock ACM.

\bibitem[\protect\citeauthoryear{Kumar \bgroup et al\mbox.\egroup
  }{2016}]{DBLP:conf/icml/KumarIOIBGZPS16}
Kumar, A.; Irsoy, O.; Ondruska, P.; Iyyer, M.; Bradbury, J.; Gulrajani, I.;
  Zhong, V.; Paulus, R.; and Socher, R.
\newblock 2016.
\newblock Ask me anything: Dynamic memory networks for natural language
  processing.
\newblock In {\em {ICML} 2016},  1378--1387.

\bibitem[\protect\citeauthoryear{Lowe \bgroup et al\mbox.\egroup
  }{2015}]{DBLP:conf/sigdial/LowePSP15}
Lowe, R.; Pow, N.; Serban, I.; and Pineau, J.
\newblock 2015.
\newblock The ubuntu dialogue corpus: {A} large dataset for research in
  unstructured multi-turn dialogue systems.
\newblock In {\em {SIGDIAL} 2015,},  285--294.

\bibitem[\protect\citeauthoryear{Lowe \bgroup et al\mbox.\egroup
  }{2017}]{DBLP:journals/dad/LowePSCLP17}
Lowe, R.~T.; Pow, N.; Serban, I.~V.; Charlin, L.; Liu, C.; and Pineau, J.
\newblock 2017.
\newblock Training end-to-end dialogue systems with the ubuntu dialogue corpus.
\newblock {\em D{\&}D} 8(1):31--65.

\bibitem[\protect\citeauthoryear{Luong \bgroup et al\mbox.\egroup
  }{2015}]{luong2015multi}
Luong, M.-T.; Le, Q.~V.; Sutskever, I.; Vinyals, O.; and Kaiser, L.
\newblock 2015.
\newblock Multi-task sequence to sequence learning.
\newblock {\em arXiv preprint arXiv:1511.06114}.

\bibitem[\protect\citeauthoryear{Miller \bgroup et al\mbox.\egroup
  }{2016}]{DBLP:journals/corr/MillerFDKBW16}
Miller, A.~H.; Fisch, A.; Dodge, J.; Karimi, A.; Bordes, A.; and Weston, J.
\newblock 2016.
\newblock Key-value memory networks for directly reading documents.
\newblock {\em CoRR} abs/1606.03126.

\bibitem[\protect\citeauthoryear{Mostafazadeh \bgroup et al\mbox.\egroup
  }{2016}]{DBLP:journals/corr/MostafazadehCHP16}
Mostafazadeh, N.; Chambers, N.; He, X.; Parikh, D.; Batra, D.; Vanderwende, L.;
  Kohli, P.; and Allen, J.~F.
\newblock 2016.
\newblock A corpus and evaluation framework for deeper understanding of
  commonsense stories.
\newblock {\em CoRR} abs/1604.01696.

\bibitem[\protect\citeauthoryear{Neelakantan \bgroup et al\mbox.\egroup
  }{2016}]{DBLP:journals/corr/NeelakantanLAMA16}
Neelakantan, A.; Le, Q.~V.; Abadi, M.; McCallum, A.; and Amodei, D.
\newblock 2016.
\newblock Learning a natural language interface with neural programmer.
\newblock {\em CoRR} abs/1611.08945.

\bibitem[\protect\citeauthoryear{Nguyen \bgroup et al\mbox.\egroup
  }{2016}]{DBLP:journals/corr/NguyenRSGTMD16}
Nguyen, T.; Rosenberg, M.; Song, X.; Gao, J.; Tiwary, S.; Majumder, R.; and
  Deng, L.
\newblock 2016.
\newblock {MS} {MARCO:} {A} human generated machine reading comprehension
  dataset.
\newblock {\em CoRR} abs/1611.09268.

\bibitem[\protect\citeauthoryear{Onishi \bgroup et al\mbox.\egroup
  }{2016}]{onishi2016did}
Onishi, T.; Wang, H.; Bansal, M.; Gimpel, K.; and McAllester, D.
\newblock 2016.
\newblock Who did what: A large-scale person-centered cloze dataset.
\newblock {\em arXiv preprint arXiv:1608.05457}.

\bibitem[\protect\citeauthoryear{Pennington, Socher, and
  Manning}{2014}]{pennington2014glove}
Pennington, J.; Socher, R.; and Manning, C.~D.
\newblock 2014.
\newblock Glove: Global vectors for word representation.
\newblock In {\em Empirical Methods in Natural Language Processing (EMNLP)},
  1532--1543.

\bibitem[\protect\citeauthoryear{Rajpurkar \bgroup et al\mbox.\egroup
  }{2016}]{rajpurkar2016squad}
Rajpurkar, P.; Zhang, J.; Lopyrev, K.; and Liang, P.
\newblock 2016.
\newblock Squad: 100,000+ questions for machine comprehension of text.
\newblock {\em arXiv preprint arXiv:1606.05250}.

\bibitem[\protect\citeauthoryear{Richardson, Burges, and
  Renshaw}{2013}]{DBLP:conf/emnlp/RichardsonBR13}
Richardson, M.; Burges, C. J.~C.; and Renshaw, E.
\newblock 2013.
\newblock Mctest: {A} challenge dataset for the open-domain machine
  comprehension of text.
\newblock In {\em {EMNLP} 2013},  193--203.

\bibitem[\protect\citeauthoryear{Ritter, Cherry, and
  Dolan}{2010}]{DBLP:conf/naacl/RitterCD10}
Ritter, A.; Cherry, C.; and Dolan, B.
\newblock 2010.
\newblock Unsupervised modeling of twitter conversations.
\newblock In {\em NAACL 2010},  172--180.

\bibitem[\protect\citeauthoryear{Serban \bgroup et al\mbox.\egroup
  }{2016a}]{Serban:2016:BED:3016387.3016435}
Serban, I.~V.; Sordoni, A.; Bengio, Y.; Courville, A.; and Pineau, J.
\newblock 2016a.
\newblock Building end-to-end dialogue systems using generative hierarchical
  neural network models.
\newblock AAAI'16,  3776--3783.
\newblock AAAI Press.

\bibitem[\protect\citeauthoryear{Serban \bgroup et al\mbox.\egroup
  }{2016b}]{DBLP:conf/acl/SerbanGGACCB16}
Serban, I.~V.; Garc{\'{\i}}a{-}Dur{\'{a}}n, A.; G{\"{u}}l{\c{c}}ehre, {\c{C}}.;
  Ahn, S.; Chandar, S.; Courville, A.~C.; and Bengio, Y.
\newblock 2016b.
\newblock Generating factoid questions with recurrent neural networks: The 30m
  factoid question-answer corpus.
\newblock In {\em Proceedings of the 54th Annual Meeting of the Association for
  Computational Linguistics, {ACL} 2016, August 7-12, 2016, Berlin, Germany,
  Volume 1: Long Papers}.

\bibitem[\protect\citeauthoryear{Serban \bgroup et al\mbox.\egroup
  }{2017}]{serban2017hierarchical}
Serban, I.~V.; Sordoni, A.; Lowe, R.; Charlin, L.; Pineau, J.; Courville,
  A.~C.; and Bengio, Y.
\newblock 2017.
\newblock A hierarchical latent variable encoder-decoder model for generating
  dialogues.
\newblock In {\em AAAI},  3295--3301.

\bibitem[\protect\citeauthoryear{Voorhees and
  Tice}{2000}]{voorhees2000building}
Voorhees, E.~M., and Tice, D.~M.
\newblock 2000.
\newblock Building a question answering test collection.
\newblock In {\em Proceedings of the 23rd annual international ACM SIGIR
  conference on Research and development in information retrieval},  200--207.
\newblock ACM.

\bibitem[\protect\citeauthoryear{Wang, Smith, and
  Mitamura}{2007}]{wang2007jeopardy}
Wang, M.; Smith, N.~A.; and Mitamura, T.
\newblock 2007.
\newblock What is the jeopardy model? a quasi-synchronous grammar for qa.
\newblock In {\em EMNLP-CoNLL}, volume~7,  22--32.

\bibitem[\protect\citeauthoryear{Yang \bgroup et al\mbox.\egroup
  }{2014}]{yang2014joint}
Yang, M.-C.; Duan, N.; Zhou, M.; and Rim, H.-C.
\newblock 2014.
\newblock Joint relational embeddings for knowledge-based question answering.
\newblock In {\em EMNLP}, volume~14,  645--650.

\bibitem[\protect\citeauthoryear{Yang, Yih, and
  Meek}{2015}]{DBLP:conf/emnlp/YangYM15}
Yang, Y.; Yih, W.; and Meek, C.
\newblock 2015.
\newblock Wikiqa: {A} challenge dataset for open-domain question answering.
\newblock In {\em {EMNLP} 2015, Lisbon, Portugal, September 17-21, 2015},
  2013--2018.

\end{thebibliography}
\bibliographystyle{aaai}

\end{document}


%
\title{\textit{Complex Sequential Question Answering}: Towards Learning to Converse Over Linked Question Answer Pairs with a Knowledge Graph: Supplementary Material}

\maketitle

\section{Example Dialogues}

~\\
\begin{enumerate}

\item

\vspace{-0.2cm}\textbf{USER:} What is name of work which is describing Percy Jackson \& the Olympians ? \\
\textbf{SYSTEM:} Camp Half-Blood chronicles \\
	
\vspace{-0.2cm}\textbf{USER:} Which collectable is a part of that work ? \\
\textbf{SYSTEM:} Percy Jackson's Greek Heroes\\

\vspace{-0.2cm}\textbf{USER:} Which collectables are a component of Camp Half-Blood chronicles or are versions of Gospel of Matthew ?\\
\textbf{SYSTEM:} Aviel revé Sant Maheu, Percy Jackson's Greek Heroes	\\
	
\vspace{-0.2cm}\textbf{USER:} And what about Framley Parsonage?\\
\textbf{SYSTEM:} Framley Parsonage	\\
	
\vspace{-0.2cm}\textbf{USER:} Which collectables are not the versions of Camp Half-Blood chronicles and were not adapted from Doctor Who ?\\
\textbf{SYSTEM:} The answer count is 54254. Do you want to see all possibilities?\\

\vspace{-0.2cm}\textbf{USER:} No, show only a few of them\\
\textbf{SYSTEM:} GURPS 4e Basic Set, The Palace of Laughter, Listening for Lions, Zen and the Brain, Fortress Besieged, The Everglades: River of Grass, Freedom from the Known, Tempest-Tost, Enchanter, Wind from the Carolinas\\
	
\vspace{-0.2cm}\textbf{USER:} Which collectables are explaining max number of works and visual arts ?	\\
\textbf{SYSTEM:} Robert Kirkman's The Walking Dead: Descent\\
	
\vspace{-0.2cm}\textbf{USER:} Does that collectable describes The Walking Dead and New Steelmaking Technology from the Bureau of Mines. Proceedings of an Open Industry Briefing Held in Association With the Electric Furnace Conference, December 8, 1987, Chicago, Il ?	\\
\textbf{SYSTEM:} YES and NO respectively\\

\\\item\vspace{0.2cm}\\

\vspace{-0.2cm}\textbf{USER:} Can you tell me what application had Total Annihilation: The Core Contingency as its original IP ?	\\
\textbf{SYSTEM:} Total Annihilation	\\
	
\vspace{-0.2cm}\textbf{USER:} Which applications were adapted from that one ?	\\
\textbf{SYSTEM:} Total Annihilation: Battle Tactics, Planetary Annihilation, Total Annihilation: The Core Contingency	\\
	
\vspace{-0.2cm}\textbf{USER:} Which applications was used as the means of distribution of around 24 applications ?	\\
\textbf{SYSTEM:} Impulse, browser game, Adobe Flash Player	\\
	
\vspace{-0.2cm}\textbf{USER:} Which works represent a fictional universe description in approximately 3 applications ?	\\
\textbf{SYSTEM:} Pirates of the Caribbean, Stargate, The Walking Dead	\\
	
\vspace{-0.2cm}\textbf{USER:} Which television programs are parts of that work ?		\\
\textbf{SYSTEM:} Did you mean Pirates of the Caribbean ?	\\
		
\vspace{-0.2cm}\textbf{USER:} No, I meant Stargate. Could you tell me the answer for that?	\\	
\textbf{SYSTEM:} Stargate Universe, Stargate Atlantis	\\
	
\vspace{-0.2cm}\textbf{USER:} Which applications are describing min number of visual arts and works ?	\\	
\textbf{SYSTEM:} Star Wars: Shadows of the Empire		\\
	
\vspace{-0.2cm}\textbf{USER:} How many applications are produced by less number of organizations and software than Vivaldi ?	\\
\textbf{SYSTEM:} 2	\\

\\\item\vspace{0.2cm}

\vspace{-0.2cm}\textbf{USER:} What is name of system software which installed Node.js ?		\\
\textbf{SYSTEM:} GNU/Linux, Solaris, Microsoft Windows	\\

\vspace{-0.2cm}\textbf{USER:} Which software is based on Node.js ?	\\
\textbf{SYSTEM:} Popcorn Time	\\
	
\vspace{-0.2cm}\textbf{USER:} Which product was based on a work from Node.js ? \\
\textbf{SYSTEM:} VPN.ht	\\

\vspace{-0.2cm}\textbf{USER:} Which programming tool is used for making that product ?	\\
\textbf{SYSTEM:} ReactJS	\\

\vspace{-0.2cm}\textbf{USER:} Which system software are that product installed on ?	\\
\textbf{SYSTEM:} Microsoft Windows, Linux	\\
		
\vspace{-0.2cm}\textbf{USER:} Which system software are installed on VPN.ht and were made for IA-64 ?	\\
\textbf{SYSTEM:} Microsoft Windows		\\
	
\vspace{-0.2cm}\textbf{USER:} How many products are the ones under which less number of system software are released than open-source software ?	\\
\textbf{SYSTEM:} two	\\
	
\vspace{-0.2cm}\textbf{USER:} How many products were adapted from Node.js or Bell 427 ?	\\
\textbf{SYSTEM:} two	\\

\\\item\vspace{0.2cm}

\vspace{-0.2cm}\textbf{USER:} Can you tell which organizations are related to BMW ?	\\
\textbf{SYSTEM:} Rolls-Royce Motor Cars, Rolls-Royce Deutschland	\\
	
\vspace{-0.2cm}\textbf{USER:} Which business enterprises are parents of those organizations ?		\\
\textbf{SYSTEM:} BMW	\\
		
\vspace{-0.2cm}\textbf{USER:} Which business enterprises were created by that one ?	\\
\textbf{SYSTEM:} Husqvarna Motorcycles, BMW Dixi	\\
	
\vspace{-0.2cm}\textbf{USER:} How many transport companies are branches of approximately the same number of business enterprises as Lima Airport Partners ?	\\
\textbf{SYSTEM:} thirty-one	\\
	
\vspace{-0.2cm}\textbf{USER:} How many social groups are subsidiaries of atmost 1 business enterprise ?		\\
\textbf{SYSTEM:} 49	\\

\vspace{-0.2cm}\textbf{USER:} How many business enterprises are parents of atleast 2 professions or transport companies ?	\\
\textbf{SYSTEM:} three	\\
	
\vspace{-0.2cm}\textbf{USER:} How many organizations or symbols are owned by BMW ?		\\
\textbf{SYSTEM:} 6	\\

\vspace{-0.2cm}\textbf{USER:} How many political territorial entities are BMW present in ?		\\
\textbf{SYSTEM:} 1	\\

\\\item\vspace{0.2cm}

\vspace{-0.2cm}\textbf{USER:} Which software are Skype installed on ?	\\
\textbf{SYSTEM:} Windows Phone, macOS, BlackBerry OS	\\

\vspace{-0.2cm}\textbf{USER:} Which system software installed Skype ?		\\
\textbf{SYSTEM:} Microsoft Windows, Symbian, Linux	\\

\vspace{-0.2cm}\textbf{USER:} Which software are installed on those system software ?	\\
\textbf{SYSTEM:} PluXml for 3rd, Yandex Browser for 1st, 2nd	\\

\vspace{-0.2cm}\textbf{USER:} Can you tell the application used for making Skype ?	\\
\textbf{SYSTEM:} Embarcadero Delphi	\\

\vspace{-0.2cm}\textbf{USER:} Which organization manufactures that service on internet ?	\\
\textbf{SYSTEM:} Skype Technologies	\\

\vspace{-0.2cm}\textbf{USER:} Which software are installed at BlackBerry OS ?	\\
\textbf{SYSTEM:} Windows Live Messenger, Google Latitude	\\

\vspace{-0.2cm}\textbf{USER:} Which software installed those ones ?	\\
\textbf{SYSTEM:} Windows 7, Windows XP, macOS	\\
	
\vspace{-0.2cm}\textbf{USER:} Which software is that one based on ?	\\
\textbf{SYSTEM:} Did you mean macOS ?	\\
	
\vspace{-0.2cm}\textbf{USER:} No, I meant Windows 7. Could you tell me the answer for that?	\\
\textbf{SYSTEM:} Windows Vista	\\
	
\vspace{-0.2cm}\textbf{USER:} Which applications are produced by less number of organizations or software than Friday the 13th: The Game ?	\\
\textbf{SYSTEM:} Pac-Man 256, Five Nights at Freddy's	\\
	
\vspace{-0.2cm}\textbf{USER:} How many system software installed Skype and OpenConcerto ?	\\
\textbf{SYSTEM:} 2		\\
	
\vspace{-0.2cm}\textbf{USER:} Which works can be used to represent a fictional universe in max number of applications ?	\\
\textbf{SYSTEM:} Shrek	\\

\vspace{-0.2cm}\textbf{USER:} Which software are launched under min number of products and software ?	\\
\textbf{SYSTEM:} Wakanda	\\

\vspace{-0.2cm}\textbf{USER:} Which software are launched under Skype and GPL-3.0 ?	\\
\textbf{SYSTEM:} Wakanda	\\

\\\item\vspace{0.2cm}

\vspace{-0.2cm}\textbf{USER:} Can you tell the administrative territory in which Louise of Savoy is present ?	\\
\textbf{SYSTEM:} France	\\
	
\vspace{-0.2cm}\textbf{USER:} Which political territories has diplomatic relation with that administrative territory ?	\\
\textbf{SYSTEM:} Australia, Germany, Austria	\\

\vspace{-0.2cm}\textbf{USER:} Which administrative territories are found in that political territory ?		\\
\textbf{SYSTEM:} Did you mean Australia ?	\\

\vspace{-0.2cm}\textbf{USER:} No, I meant Germany. Could you tell me the answer for that?	\\
\textbf{SYSTEM:} Bad Kötzting, Kappeln, Steglitz-Zehlendorf	\\

\vspace{-0.2cm}\textbf{USER:} Which administrative territories have diplomatic relations with France ?	\\
\textbf{SYSTEM:} Italy, Brazil, United States of America	\\
	
\vspace{-0.2cm}\textbf{USER:} And also tell me about Mongolia?	\\
\textbf{SYSTEM:} Russia, France, People's Republic of China	\\
	
\vspace{-0.2cm}\textbf{USER:} Which administrative territories are not related with France or are not found in People's Republic of China ?	\\
\textbf{SYSTEM:} The answer count is 31792. Do you want to see all possibilities?	\\
	
\vspace{-0.2cm}\textbf{USER:} No, show only a few of them		\\
\textbf{SYSTEM:} Ocean Township School District, Vergemoli, Penne, Abruzzo, Hundstadt, Qingjiangpu District, Carpino, Arignano, Fritzlar Principality, Duchy of Prussia, Zimmerschied	\\
	
\vspace{-0.2cm}\textbf{USER:} Can you tell which of the political territories have diplomatic relations with People's Republic of China ?		\\
\textbf{SYSTEM:} Australia, Germany	\\
	
\vspace{-0.2cm}\textbf{USER:} Can you tell me the administrative territories that border People's Republic of China ?	\\
\textbf{SYSTEM:} Russia, Japan, India	\\

\\\item\vspace{0.2cm}

\vspace{-0.2cm}\textbf{USER:} What is name of political territory in which Westerbach is found ?	\\
\textbf{SYSTEM:} Germany	\\
	
\vspace{-0.2cm}\textbf{USER:} Which watercourses are the tributaries of that one ?	\\
\textbf{SYSTEM:} Schneppenbach, Betzenbach, Hombach	\\
		
\vspace{-0.2cm}\textbf{USER:} Can you tell the watercourse used as drain for Westerbach ?		\\
\textbf{SYSTEM:} Kahl	\\
	
\vspace{-0.2cm}\textbf{USER:} How many watercourses are found at approximately nine landscapes or administrative territories ?	\\
\textbf{SYSTEM:} 5	\\
	
\vspace{-0.2cm}\textbf{USER:} Can you tell which of the administrative territories have diplomatic relations with Germany ?	\\
\textbf{SYSTEM:} Italy, Brazil, United States of America	\\
	
\vspace{-0.2cm}\textbf{USER:} Which administrative territories are diplomatically related with that one ?	\\
\textbf{SYSTEM:} Did you mean United States of America ?		\\
	
\vspace{-0.2cm}\textbf{USER:} No, I meant Italy. Could you tell me the answer for that?	\\
\textbf{SYSTEM:} Brazil, United States of America, Georgia	\\
	
\vspace{-0.2cm}\textbf{USER:} Which watercourses are a component of around one still waterss and administrative territories ?	\\
\textbf{SYSTEM:} Luisenstadt Canal, Terraced Falls, Unchalli Falls		\\
	
\vspace{-0.2cm}\textbf{USER:} Which administrative territory is that watercourse present in ?	\\
\textbf{SYSTEM:} Did you mean Luisenstadt Canal ?		\\
	
\vspace{-0.2cm}\textbf{USER:} No, I meant Unchalli Falls. Could you tell me the answer for that?	\\
\textbf{SYSTEM:} India	\\

\\\item\vspace{0.2cm}

\vspace{-0.2cm}\textbf{USER:} Which watercourses serve as Ganges 's tributaries ?	\\
\textbf{SYSTEM:} Yamuna, Ghaghara River, Kosi River	\\
	
\vspace{-0.2cm}\textbf{USER:} Which watercourses have that one as its drain ?	\\
\textbf{SYSTEM:} Did you mean Kosi River ?	\\
	
\vspace{-0.2cm}\textbf{USER:} No, I meant Yamuna. Could you tell me the answer for that?	\\
\textbf{SYSTEM:} Ken River, Tons River, Chambal River	\\
	
\vspace{-0.2cm}\textbf{USER:} Which watercourses are located in min number of political territories or administrative territories ?		\\
\textbf{SYSTEM:} Kamionka	\\
	
\vspace{-0.2cm}\textbf{USER:} Is that watercourse present at Poland and Bischofsheim ?	\\
\textbf{SYSTEM:} YES and NO respectively	\\
	
\vspace{-0.2cm}\textbf{USER:} How many watercourses are drains for Kosi River ?		\\
\textbf{SYSTEM:} one		\\
	
\vspace{-0.2cm}\textbf{USER:} And how about Santa Rosa River?	\\
\textbf{SYSTEM:} 1		\\
	
\vspace{-0.2cm}\textbf{USER:} Which administrative territories are Ganges occurring in ?	\\
\textbf{SYSTEM:} Bangladesh, India	\\

\end{enumerate}